# A Scalable Method for Solving High-Dimensional Continuous POMDPs Using Local Approximation

**Tom Erez and William D. Smart**[*]
Computer Science and Engineering
Washington University in St. Louis
St. Louis, MO 63130, USA

## Abstract

Partially-Observable Markov Decision Processes (POMDPs) are typically solved by finding an approximate global solution to a corresponding belief-MDP. In this paper, we offer a new planning algorithm for POMDPs with continuous state, action and observation spaces. Since such domains have an inherent notion of locality, we can find an approximate solution using local optimization methods. We parameterize the belief distribution as a Gaussian mixture, and use the Extended Kalman Filter (EKF) to approximate the belief update. Since the EKF is a first-order filter, we can marginalize over the observations analytically. By using feedback control and state estimation during policy execution, we recover a behavior that is effectively conditioned on incoming observations despite the unconditioned planning. Local optimization provides no guarantees of global optimality, but it allows us to tackle domains that are at least an order of magnitude larger than the current state-of-the-art. We demonstrate the scalability of our algorithm by considering a simulated hand-eye coordination domain with 16 continuous state dimensions and 6 continuous action dimensions.

## 1 INTRODUCTION

Partially-Observable Markov Decision Processes (POMDPs) offer a framework for studying decision making under uncertainty. The optimal behavior in a POMDP domain is expected to strike a balance between exploring the partially-observable world and acting in a goal-directed manner. Most of the POMDP literature is concerned with discrete domains, but in the past few years, as POMDP tools become more powerful, there is growing interest in tackling continuous domains (Porta et al., 2006; Brooks, 2009).

The standard approach to solving POMDPs is to find an approximate solution to the fully-observable *belief*-MDP, whose states are probability distributions over the state space of the original POMDP (Kaelbling et al., 1998). In the discrete case, the resulting belief space is continuous but finite-dimensional, and belief update can be carried out exactly. However, the belief space of a continuous POMDP is infinite-dimensional, and must be approximated (Thrun, 2000).

The optimal value function of belief-MDPs is piecewise-linear and convex in the discrete case (Sondik, 1971), and this also holds for some cases of continuous state (Porta et al., 2006), as long as the observations and actions are discrete. This result was used to tackle domains with continuous hybrid-linear dynamics by Brunskill et al. (2008). Other combinations of the discrete and the continuous domains were also considered (Hoey & Poupart, 2005; Spaan & Vlassis, 2005). The richest domain tackled by continuous POMDPs is probably outdoor navigation (Brooks, 2009).

However, in all the examples mentioned above, the belief domain is solved through *global* optimization. Since the volume of state space grows exponentially with the dimension of the state, it is unrealistic to seek a globally-optimal solution in domains above a certain size because of the *curse of dimensionality*. Some studies (e.g., Feng & Zilberstein, 2004) try to offset some of the computational burden by finding parts of belief space that can safely be ignored, but the fundamental problem of exponential scaling remains.

In contrast, continuous domains naturally admit a notion of distance, which allows the application of *local* optimization methods. Here, we present a method for approximating a locally-optimal solution to a POMDP in which state, action and observation space are continuous. This work is a departure from the current POMDP literature, as it offers a different trade-off between provable correctness and scalability. Since we employ a local method, guarantees or bounds for global optimality are impossible to obtain. However, local optimization is not subject to the curse of dimensionality, and can tackle domains that are outside the reach of global approaches.

---

[*] {etom,wds}@cse.wustl.edu

In this paper, we use Differential Dynamic Programming (DDP) to solve for a locally-optimal policy (section 4). While DDP optimizes the open-loop ("blind") policy, it also outputs a feedback policy (section 5) that is used to adapt the execution-time behavior to the actual incoming observations.

We approximate the belief space with a parametric distribution, specifically a Gaussian mixture, and use the Extended Kalman Filter (EKF) for belief update (Stengel, 1994). By virtue of the EKF being a first-order filter, we can analytically marginalize the belief update over the observations. By focusing only on the most-likely observation, we recover a deterministic update scheme (section 3). This seems counter-intuitive, since the goal of solving POMDPs is to generate behavior that responds to observations. However, note that observations are marginalized only for planning; during policy execution, the actual incoming observations are taken into account as they are used to estimate the agent's hidden state. By coupling state estimation and feedback control, the agent's behavior is again conditioned on incoming observations, allowing it to respond to the changing environment in real time.

POMDPs are often used to tackle domains with unilateral constraints, such as contacts (e.g., Hsiao et al., 2007). Since the EKF works by linearizing the dynamics, a single Gaussian would not be descriptive enough to handle such discontinuities. Since the distribution of the hidden state is truncated by a constraint manifold, we explicitly approximate the probability mass that aggregates on this manifold with a Gaussian of lower rank (section 3.2). We analytically account for the flow of probability mass between the two Gaussians using the equations of truncated normal distributions (section 3.2.1). While these approximations are used for belief propagation during planning, more accurate state estimation (e.g., a particle filter) can be employed during policy execution (section 5).

The scalability of the proposed method is unmatched by any existing technique, and allows the use of POMDPs in application domains that are too large to admit global solutions. In section 6.2, we apply our method to a simulated domain of hand-eye coordination with 16 continuous state dimensions and 6 continuous action dimensions.

The principles of deterministic planning through marginalized observations is not new: for example, Prentice & Roy (2009) also employ a single-Gaussian approximation of the belief state during planning. However, this approach requires samples that span the entire state space, and is hence bound by the curse of dimensionality. The notion of marginalizing the observations during local optimization has informed the solution method presented in Erez & Smart (2009). The method we present here is very similar to Miller et al. (2009), who used the term *nominal-belief optimization*, and Platt et al. (2010), who used the term *maximum-likelihood observations*. However, neither approaches can tackle domains with unilateral constraints (section 3.2).

## 2 DEFINITIONS

We consider a discrete-time POMDP defined by a tuple $\langle S, A, Z, T, \Omega, R, N \rangle$, where: $S, A$ and $Z$ are the state space, action space and observation space, respectively; $T(s', s, a) = \Pr(s'|s, a)$ is a transition function describing the probability of the next state given the current state and action; $\Omega(z, s, a) = \Pr(z|s, a)$ is the observation function, describing the probability of an observation given the current state and action; and R is a time-dependent reward function $R^i(s, a)$, with a terminal reward $R^N(s)$. In this paper we consider an undiscounted optimality criterion, where the agent's goal is to maximize the expected cumulative reward within a fixed time horizon $N$. This formulation is a deviation from the common focus on discounted horizons, and we adopt it because it is useful for the local optimal control algorithm we employ (section 4).

## 3 THE BELIEF DOMAIN

The *belief state* $b \in B$ is a probability distribution over $S$, where $b^i(s)$ is the likelihood of the true state being $s$ at time $i$. In order to construct the belief domain of a given POMDP, we need to find a representation for $b$, and define the reward function and dynamics (belief update) over this space.

Given the current belief $b$, an action $a$ and observation $z$, the updated belief $b'$ can be calculated by applying Bayes's rule. In the continuous case $B$ is infinite-dimensional, and Bayes's rule yields an integral:

$$b'(s') \propto \int_s b(s) T(s', s, a) \Omega(z, s, a) ds.$$

In order to make this function computationally tractable, we must employ some approximation $\hat{b}$ to the true belief $b$, and commit to some state estimation filter to update the approximated belief.

The reward associated with a belief is simply the expected value over this state distribution:

$$R^i(b, a) = \mathop{\mathrm{E}}_{s \sim b} \left[ R^i(s, a) \right]. \qquad (1)$$

Since our optimality criterion employs a finite-horizon, our optimization focuses on the *time-dependent policy* $\pi(\hat{b}, i)$,

mapping beliefs and time to actions. The optimal policy maximizes the cumulative reward:

$$\pi^* = \underset{\pi}{\arg\max}\, \mathrm{E}\Big[\sum_{i=1}^{N} R^i(b^i, \pi(b^i, i))\Big]. \quad (2)$$

## 3.1 CONTINUOUS DYNAMICS

In this section, we focus on nonlinear stochastic dynamics $ds = f(s,a)dt + q(s,a)d\xi$, where $\xi$ is a Wiener process. For a given state $s$ and action $a$, integrating these dynamics over a small time-step $\tau$ results in a normal distribution over the next state $s'$:

$$T(s', s, a) = \mathcal{N}(s'|F(s,a), Q(s,a)),$$

where the mean is propagated with the Euler integration

$$F = s + \tau f(s, a), \quad (3)$$

and the covariance $Q = \tau q^\mathsf{T} q$ is a time-scaling of the continuous process $qd\xi$.

Similarly, we focus on observation distributions of the form: $\Omega(z, s, a) = \mathcal{N}(z|w(s), W(s,a))$, where $w$ deterministically maps states to observations, and $W$ describes how the current state and action affect the observation noise.

Given a Gaussian prior on the initial state, we approximate the infinite-dimensional $b$ by a single Gaussian:

$$\hat{b}(s) = \mathcal{N}(s|\hat{s}, \Sigma) = \frac{1}{(2\pi)^{\frac{k}{2}}|\Sigma|^{\frac{1}{2}}} e^{-\frac{1}{2}(s-\hat{s})^\mathsf{T} \Sigma^{-1}(s-\hat{s})},$$

and denote its parameterization by:

$$\nu = \{\hat{s}, \Sigma\} \quad (4)$$

where the covariance $\Sigma$ belongs to the space of symmetric, positive-semidefinite matrices $\mathcal{M} \subset \mathbb{R}^{n \times n}$. Therefore, the belief space $\hat{B}$ is parameterized in this case by the product space $\nu \in S \times \mathcal{M}$. In the limit of $\tau \to 0$, this approximation is accurate.

In order to approximate the belief update, we use the Extended Kalman Filter (EKF) (Stengel, 1994). Given the current belief $\hat{b}$, action $a$ and observation $z$, we calculate the partial derivatives around $\hat{s}$: $w_s = \partial w/\partial s$ and $F_s = \partial F/\partial s$. We find the uncorrected estimation uncertainty $H = F_s \Sigma F_s^\mathsf{T} + Q(\hat{s}, a)$ and calculate the new mean $\hat{s}'$ by the innovation process:

$$\hat{s}' = F(\hat{s}, a) - K(z - w(\hat{s})). \quad (5)$$

where $K = H w_s (w_s^\mathsf{T} H w_s + W(\hat{s}, a))^{-1}$ is the *Kalman gain*. Finally, the new covariance $\Sigma'$ is given by:

$$\Sigma' = \Psi(\hat{s}, \Sigma, a) =$$
$$H - H w_s (w_s^\mathsf{T} H w_s + W(\hat{s}, a))^{-1} w_s^\mathsf{T} H^\mathsf{T}. \quad (6)$$

The deterministic belief update is obtained by marginalizing equations (5) and (6) over the observation $z$. Equation (5) is linear in $z$, and so we can take the expectation by simply replacing $z$ with its mean $w(\hat{s})$. The second term of equation (5) vanishes, and so the mean follows (3). By virtue of the EKF being a first-order filter, the calculation in (6) is independent of $z$. In summary, the deterministic belief update is formed by the combination of (3) and (6):

$$\hat{b}' = \{F(\hat{s}, a), \Psi(\hat{s}, \Sigma, a)\}. \quad (7)$$

## 3.2 UNILATERAL CONSTRAINTS

In the previous section, we made the assumption that $F$ and $w$ can be linearized WRT $s$. However, this assumption may be too restrictive for some domains; in particular, it excludes discontinuous dynamics that occur due to unilateral constraints. Since this category includes interesting domains of disambiguation by contact, object manipulation and locomotion, we extend our method to handle the non-Gaussian beliefs that come about in such cases.

In this section we consider domains with non-penetration constraints $\Gamma$:

$$\begin{aligned} ds &= f(s,a)dt + Q(s,a)d\xi, \\ \Gamma(s) &\geq 0. \end{aligned} \quad (8)$$

In the general case, the reaction forces that enforce these constraints can be calculated using complementarity methods (Stewart, 2000) or penalty methods (Drumwright, 2008). When $\Gamma(s) = 0$, we say that the constraint is *active*. In this paper, we consider domains where at most one constraint is active at any one time, and so we may focus on cases where $\Gamma(s)$ is scalar.

The resulting belief $b$ can no longer be described by a simple normal distribution: $\Gamma$ describes an $(n-1)$-dimensional constraint manifold, and the belief distribution is truncated at this manifold, with some probability mass aggregating on it. We approximate this truncated distribution with a weighted mixture of two Gaussians: one describing the belief distribution in the unconstrained volume, and the other describing the aggregated belief on the constraint (hence degenerate in the direction locally perpendicular to the manifold). Using $\nu$ to parameterize a single Gaussian as in (4), we denote the parameterized belief

$$\hat{b}(s) = \alpha \mathcal{N}(s|\hat{s}_1, \Sigma_1) + (1-\alpha) \mathcal{N}(s|\hat{s}_2, \Sigma_2)$$

by the shorthand

$$\hat{b} = \{\nu_1, \nu_2, \alpha\},$$

where $\alpha \in [0, 1]$ is the relative weight of the first Gaussian. This is not an exact representation of the true belief; a Gaussian has infinite support, and therefore the unconstrained Gaussian has non-zero probability mass beyond the constraint. However, this mass is small enough that, in practice, it has had no noticeable effect on our results.

Belief update is done in two stages, as outlined in algorithm 1. In the first stage, we update the belief of each Gaussian independently using (7). Assuming that there is noise in the direction locally-perpendicular to the constraint, the second Gaussian is now full-rank. In the second stage, we re-approximate this two-Gaussian mixture, ensuring that the resulting mixture maintains the form described above — the probability mass above the constraint manifold is approximated with one Gaussian, and the belief that lies below the constraint is approximated with a second, degenerate Gaussian that lies on the manifold. The details of the computations required for the second stage are detailed in the next two subsections.

### 3.2.1 Truncation

In order to re-adjust the belief to the constraint, we linearize the constraint function $\Gamma \approx Js + e \geq 0$ around the mean of each Gaussian. We compute the distributions on either side of the constraint analytically by considering truncated normal distributions (Boutilier, 2002; Toussaint, 2009). We can linearly rotate and re-scale the state space so as to ensure that the constraint manifold is locally perpendicular to the $k^{\text{th}}$ dimension of $s$, and that the uncertainty in this dimension is independent of the others. Therefore, we can focus our analysis on the one-dimensional case, assuming without loss of generality that the constraint does not affect any dimension but $k$.

Let $x \sim \mathcal{N}(\mu, \sigma^2)$. When bound to an interval $x \in [l, u]$, its distribution becomes:

$$Pr(x) \propto \frac{1}{\sqrt{2\pi\sigma^2}} \exp\left(-\frac{(x-\mu)^2}{2\sigma^2}\right) \Theta(x-l)\Theta(u-x),$$

where $\Theta$ is the Heaviside function. The first two moments of the resulting distribution are:

$$\mathrm{E}(X \mid l < X < u) = \mu + \sigma \frac{\phi(\bar{l}) - \phi(\bar{u})}{\Phi(\bar{u}) - \Phi(\bar{l})} \quad (9a)$$

$$\mathrm{Var}(X \mid l < X < u) =$$
$$\sigma^2 \left[1 + \frac{\bar{l}\phi(\bar{l}) - \bar{u}\phi(\bar{u})}{\Phi(\bar{u}) - \Phi(\bar{l})} - \left(\frac{\phi(\bar{l}) - \phi(\bar{u})}{\Phi(\bar{u}) - \Phi(\bar{l})}\right)^2\right] \quad (9b)$$

where $\bar{l} = \frac{l-\mu}{\sigma}, \bar{u} = \frac{u-\mu}{\sigma}$, and $\phi(\bar{x})$, $\Phi(\bar{x})$ are the PDF and CDF of the normal distribution with zero mean and unit variance. The probability masses that aggregate on the constraints are $\Phi(\bar{l})$ and $1 - \Phi(\bar{u})$. We are interested in distributions over one-sided intervals, so either $l = -\infty$ or $u = \infty$, which further simplifies (9).

**Algorithm 1** Deterministic Belief Update with Unilateral Constraints

**Input:** $\hat{b} = \{\nu_1, \nu_2, \alpha\}$, action $a$
**for** $i = 1, 2$ **do**
  **Marginalized EKF:** Calculate $\nu_i'$ by (7).
  **Truncation:** Calculate $\{\nu_i^u, \nu_i^l, \alpha_i^u\}$ by (9).
**end for**
**Reduction:** Calculate $\nu_1'', \nu_2''$ by (10).
**Adjustment:** Project $\nu_2''$ onto constraint by (11).
**Weight update:** Calculate $\alpha'$ by (12).
**Output:** $\hat{b}' = \{\nu_1'', \nu_2'', \alpha'\}$.

### 3.2.2 Mixture Reduction

We use the truncation procedure described above to split each Gaussian in two, across the constraint. In order to maintain our form (one Gaussian unconstrained, one Gaussian on the constraint manifold), we reduce this four-Gaussian mixture back to two, and project the second Gaussian onto the constraint.

Reducing a mixture of two Gaussians $\{\nu_1, \nu_2, \alpha\}$ results in a single Gaussian whose mean $\hat{s}$ and covariance $\Sigma$ are:

$$\hat{s} = \alpha\hat{s}_1 + (1-\alpha)\hat{s}_2, \quad (10a)$$

$$\Sigma = \alpha\Sigma_1 + (1-\alpha)\Sigma_2 + \alpha(1-\alpha)(\hat{s}_1-\hat{s}_2)(\hat{s}_1-\hat{s}_2)^{\mathsf{T}} \quad (10b)$$

Using these equations, we combine the two Gaussians above the constraint into a single Gaussian $\nu_1''$, and the two Gaussians below the constraint into $\nu_2''$. Assuming that the constraint is locally perpendicular to the $k^{th}$ dimension as above, we project $\nu_2''$ onto the constraint by setting:

$$(\hat{s}_2'')_k = \Gamma(\hat{s}_2''), \text{ and } (\Sigma_2'')_{k,k} = 0. \quad (11)$$

Finally, the weight of the unconstrained Gaussian in the adjusted mixture is:

$$\alpha' = \alpha\alpha_1^u + (1-\alpha)\alpha_2^u. \quad (12)$$

## 4 PLANNING

The belief update schemes of the previous section (together with (1)) define a problem of deterministic optimal control in a high-dimensional continuous space, with non-linear dynamics and non-quadratic reward. This can be solved using a variety of techniques; Platt et al. (2010), for example, use the *multiple shooting* method. We use a local optimization scheme called Differential Dynamic Programming (DDP), an algorithm that has been successfully applied to real-world high-dimensional, non-linear control domains (e.g., Abbeel & Ng, 2005; Tassa et al., 2008). The interested reader may find an in-depth description of the algorithm in Jacobson & Mayne (1970).

The benefit of using DDP is that in addition to the locally-optimal trajectory and the open-loop action sequence which realizes this trajectory, it outputs a sequence of linear feedback gain matrices. These parameterize the policy (section 5) to create a feedback controller for the original POMDP.

## 5 POLICY EXECUTION

Since a policy for a continuous POMDP is infinite-dimensional, it also needs to be parameterized. In this paper we focus on policies that are locally-linear:

$$\pi(\hat{b}, i) = \bar{a}^i + L^i(\hat{b} - \bar{b}^i) \quad (13)$$

for some parameterized belief states $\bar{b}^{1:N}$, actions $\bar{a}^{1:N-1}$ and feedback gain matrices $L^{1:N-1}$. This parameterization corresponds to the output of Differential Dynamic Programming, as described in the previous section.

The policy is executed post-planning, as the agent interacts with the environment. Incoming observations are filtered by state estimation, and feedback control responds to changes in the perceived state and reacts appropriately. Thus, the agent's behavior is conditioned on received observations even though these were marginalized during planning.

At this stage, we are no longer committed to the belief update schemes of section 3, and a more accurate filter (e.g., particle filter) can be used for state approximation. This new filter may employ a different representation of the belief state which may be richer than a single Gaussian.

## 6 RESULTS

First, we demonstrate key features of our method by considering an example of planar navigation, roughly corresponding to domains considered by Roy & Thrun (1999) and Brooks (2009). Then, we demonstrate the scalability of our method by solving a 16-dimensional problem, first presented by Erez & Smart (2009).

### 6.1 PLANAR NAVIGATION

In this problem, a robot must move in a closed room from a start point to a target while avoiding obstacles. The robot cannot sense its position, but may localize itself by making contact with the walls. Here, state, action and observation are all two-dimensional, and the constraint is scalar. The resulting optimal behavior (figure 1(a)) is found in less than a minute: the robot avoids the obstacles by approaching the side wall, and then "cut" the corner on its way to the target at the target at the bottom wall. In order to study the effect of linearizing the constraint, we tested a case where the agent interacts with the curved segment of the constraint.

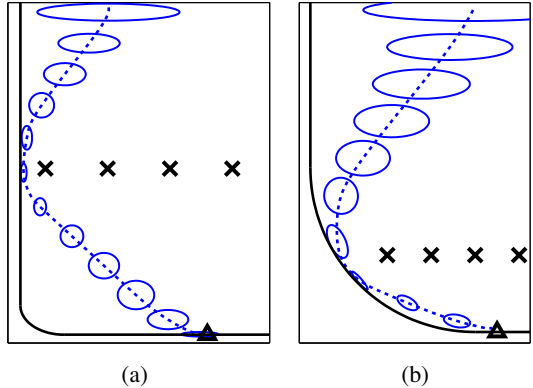

Figure 1: 2D navigation. **(a)** The robot (dashed blue line) localizes itself by approaching the wall (black) while avoiding the obstacles (X) and reaching the target (triangle). Blue ellipses depict the covariance. **(b)** An optimal solution that interacts with the curved part of the constraint manifold.

As figure 1(b) shows, the optimal path in this case follows the round corner without difficulty.

The disambiguating property of the contact with the wall is termed "coastal navigation" (Roy & Thrun, 1999), and our algorithm is able to identify and leverage this feature as it emerges in the optimal solution. We cannot offer a direct comparison of our results with Brooks (2009), since his experiments were conducted on real robots. Note the qualitative difference in our approach: Brooks's method approximates the global optimum, and requires 8000 samples that are processed in ~25 minutes, while our method finds a locally-optimal solution in less than one minute.

### 6.2 HAND-EYE COORDINATION

This problem illustrates the scalability of our algorithm, since we believe it cannot be solved by any other POMDP technique. In addition, we demonstrate reactive behavior through feedback control. The resulting behavior is best illustrated by the supplementary movie[1].

This domain is similar to the one we solved in Erez & Smart (2009), with few notable differences: first, here we introduce perceptual inhibition during saccadic motion; more significantly, the solution method of this paper is faster and more robust than the minimax approach presented there, since here we do not solve for the adversarial behavior.

This domain simulates the problem of an agent coordinating two "hands" and an "eye". The task requires the agent to bring the hands from their starting positions to a target point at a specific time, while avoiding four obstacles in a planar scene. State transitions are subject to a fixed Gaussian process noise. The obstacles' positions are unknown,

---

[1] www.cse.wustl.edu/~etom/uai2010.mp4

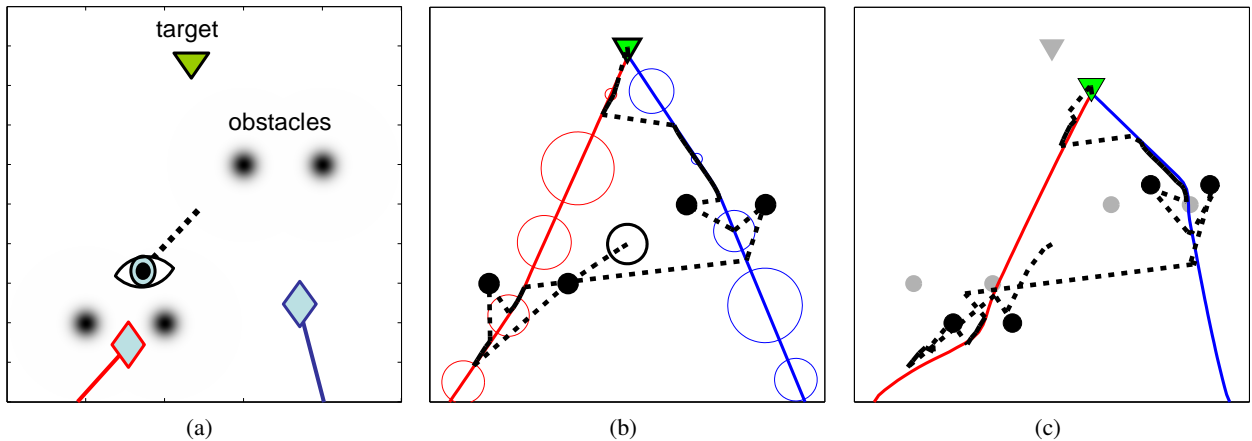

Figure 2: Hand-eye coordination. **(a)** A schematic of the scene. The hands (red and blue diamonds) aim for the target (green triangle) while avoiding the obstacles whose position is uncertain (black blurs), assisted by the eye (middle). **(b)** An illustration of the learned optimal trajectory. The size of the eye's fovea is represented as a black circle around its starting place (center of the figure). The eye's trajectory alternates between following the hands in smooth pursuit (thick black line) and saccadic motion (thin dashed line) between the hands and the obstacles. The uncertainty in the hands' positions are depicted as a series of circles – note that when the eye is smoothly following a hand, the uncertainty vanishes. **(c)** If the positions of obstacles and target change from those during training (grey), feedback control and state estimation allow the agent to adjust its behavior. These behaviors are best illustrated by the supplementary movie[1].

so the agent must observe and estimate these as well.

The planar scene is illustrated in figure 2(a). The state is defined in terms of the following variables: $s_e$ is the eye's two-dimensional position, $s_{h_1}$ and $s_{h_2}$ are the positions of the hands, $s_t$ is the target's position, and $\{s_{l_i}, i = 1 \ldots 4\}$ are the positions of four obstacles. Therefore, the state space has 16 continuous dimensions. Every state $s$ is a concatenation of the 8 planar positions above. The action space $A$ is 6-dimensional, specifying planar velocities for the hands and eye. As stated in the previous section, such a domain is infeasible for global optimization, and cannot be solved by any existing global POMDP algorithm.

$Z$, the observation space, is identical to the state space. The observation noise covariance $W$ is diagonal, allowing independent observation of each scene element. $W$ is state- and action-dependent: the eye has the capacity to produce unambiguous observations in a small region around its current position, conceptually modelling foveated vision. The eye's gaze locally reduces the observation noise:

$$W_\star(s, a) = 1 - e^{-\|s_e - s_\star\|^2/2\eta} + 0.01 a_e^\mathsf{T} a_e \qquad (14)$$

where $\star$ stands for one of the scene elements: $h_1, h_2, t$, or any of the obstacles $l_i$. The parameter $\eta$ determines the size of the fovea, and $a_e$ is the current actuation of the eye. The last RHS term in (14) models visual inhibition during saccadic eye movement, effectively eliminating the eye's disambiguating effect during high-velocity eye movements. Thus, the eye produces valid observations only when it is close to an object, and moving slowly.

The reward function is the same as in Erez & Smart (2009) — it penalizes for distance between the hands and the target at the final time step, and for proximity between the hands and the obstacles at all other time steps; action (displacement of hands and eye) incurs a quadratic cost, where the eye's action cost is negligibly small.

The covariance of the process noise $Q$ is a constant diagonal matrix, where the noise in the X- and Y-direction are equal for every scene element. The process noise that affects the eye, obstacles and target is zero. From the agent's perspective, this means that once observed, the positions of the target and obstacles can be trusted to remain unmoved, allowing the eye's position to provide grounding for locating all other elements of the scene. Since process noise is uncorrelated between state dimensions and symmetric in both planar directions, the belief covariance can be decomposed and succinctly represented by a single scalar variance for each scene element (but the eye, whose position is certain). In all, $\hat{B}$ has 23 dimensions.

Figure 2(b) shows the resulting locally-optimal trajectories for both hands and the eye. Notice how the eye tracks each hand in turn as it passes close the obstacles, and how the hands time their approach to the obstacles to synchronize with the eye. Interestingly, in the optimal solution we can see the emergence of two distinct phases of eye behavior – smooth pursuit, where the eye tracks the hand, and saccades, where the eye rapidly moves from one gaze target to another. This behavior is in accordance with biological visual behavior (Cassin & Rubin, 2001).

To demonstrate the responsiveness of the resulting policy, we tested the agent's feedback control in a modified scene, where the obstacles were shifted from their position during planning. Figure 2(c) shows the resulting behavior: the hands' trajectories are adjusted as the eye perceives and updates the estimated position of the obstacles and hands (using EKF). Since feedback is specified over belief space, the behaving agent also responds to changes in the estimation uncertainty.

In our initial experiments, we observed mis-convergence to a local minimum, where the eye would not bother to saccade between the two hands, instead sticking to only one of them. This was remedied by employing a shaping-continuation method (Erez & Smart, 2008) We first found a solution to a simpler problem, where the size of the notional fovea is large ($\eta = 10$). There, the wide field-of-view allowed a relatively unambiguous view of the entire scene, and enabled the formation of a trajectory that had the right coarse features (a move to the left, then a move to the right, then a move up), even as it was not required to perform precise saccades. As learning progresses, the size of the foveal region was gradually reduced, making exact eye movements more important. Every new problem instance was solved using the previous solution as a starting point. This process repeated for decreasing fovea radius ($\eta = [1, 0.3, 0.05]$) until we generated a solution to the original problem. The shaping sequence required running DDP to convergence 4 times, yet the optimal solution for this 16-dimensional domain was found in less than 3 minutes of MATLAB running on a single-core desktop computer.

## 7 DISCUSSION

This paper offers a new perspective on solving continuous POMDPs. Instead of using global approximation in a belief-MDP, we marginalize the observations and cast the infinite-dimensional, stochastic belief domain in terms of a finite-dimensional optimal control problem. The approach we present here applies to domains whose dynamics are smooth (and hence Gaussian beliefs constitute a good approximation) or with one unilateral constraint (where we use a specifically-structured two-Gaussian mixture); however, our approximations may not hold for domains with more elaborate structure.

While this method scales very well with state dimensionality, we chose to focus on domains where only one constraint is active at a time. Such cases are amenable to analytic manipulation using truncated normal distributions, as described above. If we extended this type of analysis to cases where more than one constraint may be active at once, we would be assigning a Gaussian to every combination of active constraints, and accounting for the flow of probability mass between all of them. This would introduce yet another set of approximations, and would be computationally reasonable only for a small number of jointly-active constraints.

One natural extension of this work could employ local optimization from multiple starting points, creating a controller that uses a trajectory library (Stolle & Atkeson, 2006; Tassa et al., 2008). Such a scheme could extend the basin of attraction of our local controller (similar to Tedrake, 2009), and produce a better approximation of the globally-optimal policy. In particular, a multi-modal prior can be handled by finding the optimal behavior for each of the modes, and using state estimation during policy execution to choose the relevant case. If behavior is expected to encounter different variants of the same environment, which vary along a single parameter (e.g. ground incline, or object mass), we can extend a single solution and create a manifold controller (Erez & Smart, 2007) through some continuation approach.

In many real-life cases, an active constraint results in frictional forces, in addition to the reaction forces that maintain non-penetration. This can be incorporated into our method by using a different dynamical model for the initial belief update of the constrained Gaussian $\nu_2$, in particular one that incorporates friction. In cases where making contact (i.e., collision) is associated with a non-negligible impact dynamics of other degrees of freedom beyond the constrained one (e.g., foot-ground impact, or ball-racket impact), these impulses can be considered as we project the Gaussian that lies below the constraint manifold onto the linearized hyperplane.

## Acknowledgements

This work was supported by NSF award BCS 0924609

## References

Abbeel, Pieter and Ng, Andrew Y. Exploration and apprenticeship learning in reinforcement learning. In *International Conference on Machine Learning (ICML)*, pp. 1–8, 2005.

Boutilier, Craig. A POMDP formulation of preference elicitation problems. In *Eighteenth national conference on Artificial intelligence*, pp. 239–246, 2002.

Brooks, Alex. *Parametric POMDPs*. VDM Verlag, 2009.

Brunskill, Emma, Kaelbling, Leslie, Lozano-Perez, Tomas, and Roy, Nicholas. Continuous-state POMDPs with hybrid dynamics. In *Tenth International Symposium on Artificial Intelligence and Mathematics (ISAIM)*, January 2008.

Cassin, Barbara and Rubin, Melvin L. *Dictionary of Eye Terminology*. Triad Publishing Company (FL), 2001.

Drumwright, E. A fast and stable penalty method for rigid body simulation. *IEEE Transactions on Visualization and Computer Graphics*, 14(1):231–240, 2008.

Erez, T. and Smart, W.D. Bipedal walking on rough terrain using manifold control. In *IEEE/RSJ International Conference on Intelligent Robots and Systems (IROS)*, pp. 1539–1544, 2007.

Erez, Tom and Smart, William D. What does shaping mean for computational reinforcement learning? *7th IEEE International Conference on Development and Learning (ICDL)*, pp. 215–219, 2008.

Erez, Tom and Smart, William D. Coupling perception and action using minimax optimal control. In *IEEE Symposium on Adaptive Dynamic Programming and Reinforcement Learning (ADPRL)*, pp. 58–65, 2009.

Feng, Zhengzhu and Zilberstein, Shlomo. Region-based incremental pruning for POMDPs. In *The 20th conference on Uncertainty in artificial intelligence (UAI)*, pp. 146–153, 2004.

Hoey, Jesse and Poupart, Pascal. Solving POMDPs with continuous or large discrete observation spaces. In *International Joint Conference on Artificial Intelligence (IJCAI)*, pp. 1332–1338, 2005.

Hsiao, Kaijen, Kaelbling, Leslie Pack, and Lozano-Perez, Tomas. Grasping POMDPs. In *IEEE International Conference on Robotics and Automation (ICRA)*, pp. 4685–4692, 2007.

Jacobson, D. H. and Mayne, D. Q. *Differential Dynamic Programming*. Elsevier, 1970.

Kaelbling, Leslie Pack, Littman, Michael L., and Cassandra, Anthony R. Planning and acting in partially observable stochastic domains. *Artificial Intelligence*, 101(1-2):99–134, 1998.

Miller, Scott A., Harris, Zachary A., and Chong, Edwin K. P. A POMDP framework for coordinated guidance of autonomous uavs for multitarget tracking. *EURASIP Journal of Advanced Signal Process*, 2009:1–17, 2009.

Platt, Robert, Tedrake, Russ, Kaelbling, Leslie Pack, and Lozano-Perez, Tomas. Belief space planning assuming maximum likelihood observations. In *Robotics: Science and Systems (R:SS)*, 2010.

Porta, Josep M., Vlassis, Nikos, Spaan, Matthijs T.J., and Poupart, Pascal. Point-based value iteration for continuous POMDPs. *Journal of Machine Learning Research*, 7:2329–2367, December 2006.

Prentice, S. and Roy, N. The belief roadmap: Efficient planning in belief space by factoring the covariance. *The International Journal of Robotics Research*, 28(11-12):1448–1465, 2009.

Roy, Nicholas and Thrun, Sebastian. Coastal navigation with mobile robots. In *Advances in Neural Processing Systems (NIPS)*, volume 12, pp. 1043–1049, 1999.

Sondik, E.J. *The Optimal Control of Partially Observable Markov Processes*. PhD thesis, Stanford, 1971.

Spaan, Matthijs T. J. and Vlassis, Nikos A. Planning with continuous actions in partially observable environments. In *IEEE International Conference on Robotics and Automation (ICRA)*, pp. 3458–3463, 2005.

Stengel, Robert F. *Optimal Control and Estimation*. Dover Publications, 1994.

Stewart, David E. Rigid-body dynamics with friction and impact. *SIAM Reviews*, 42(1):3–39, 2000.

Stolle, Martin and Atkeson, Chris. Policies based on trajectory libraries. In *IEEE International Conference on Robotics and Automation (ICRA)*, 2006.

Tassa, Yuval, Erez, Tom, and Smart, William. Receding horizon differential dynamic programming. In *Advances in Neural Information Processing Systems (NIPS)*, volume 20. MIT Press, Cambridge, MA, 2008.

Tedrake, Russ. LQR-trees: Feedback motion planning on sparse randomized trees. In *Robotics: Science and Systems (R:SS)*, 2009.

Thrun, S. Monte carlo POMDPs. In Solla, S.A., Leen, T.K., and Müller, K.-R. (eds.), *Advances in Neural Information Processing Systems (NIPS)*, volume 12, pp. 1064–1070. MIT Press, 2000.

Toussaint, Marc. Pros and cons of truncated gaussian ep in the context of approximate inference control. NIPS workshop on Probabilistic Approaches for Robotics and Control, 2009.